\RequirePackage{amsmath}
\documentclass{llncs}

\usepackage{graphicx}
\usepackage{algorithm}
\usepackage{algpseudocode}
\usepackage[verbose]{wrapfig}
\usepackage[caption=false]{subfig}

\usepackage{amsmath}
\usepackage{amssymb}
\usepackage{amsfonts}

\usepackage{xifthen}

\usepackage{xcolor}
\newcommand{\dale}{Dale M$^c$Conachie}

\DeclareMathOperator*{\argmin}{argmin}
\DeclareMathOperator*{\argmax}{argmax}
\DeclareMathOperator*{\diag}{diag}
\DeclareMathOperator*{\proj}{Proj}
\newcommand{\numberthis}{\addtocounter{equation}{1}\tag{\theequation}}

\usepackage{mathtools}
\DeclarePairedDelimiter{\ceil}{\lceil}{\rceil}

\newcommand{\reals}{\mathbb{R}}

\newcommand{\se}[1]{SE(#1)}
\newcommand{\tanse}[1]{\mathfrak{se}(#1)}
\newcommand{\eye}{\mathbf{I}}
\newcommand{\normal}[2]{\mathcal{N}\left(#1,#2\right)}

\newcommand{\numgrippers}{G}
\newcommand{\gripperindex}{g}
\newcommand{\gripperconfig}{q}
\newcommand{\grippervelocity}{\dot \gripperconfig}
\newcommand{\robotconfig}{\gripperconfig}
\newcommand{\robotvelocity}{\dot \robotconfig}

\newcommand{\robotconfigspace}{\se3^\numgrippers}
\newcommand{\robotvelocityspace}{\tanse3^\numgrippers}

\newcommand{\maxgrippervel}[1][]{\grippervelocity_%
    {\ifthenelse{\isempty{#1}}{\textrm{max}}{\text{max},#1}}
}
\newcommand{\maxgrippervelservo}{\maxgrippervel[e]}
\newcommand{\maxgrippervelobstacle}{\maxgrippervel[o]}

\newcommand{\relaxeddistancematrix}{D}
\newcommand{\deformconfig}{\mathcal{P}}
\newcommand{\deformvelocity}{\dot \deformconfig}
\newcommand{\numdeformpoints}{P}
\newcommand{\deformconfigspacesize}{{3\numdeformpoints}}
\newcommand{\deformconfigspace}{\reals^\deformconfigspacesize}

\newcommand{\deformtarget}{\mathcal{T}}
\newcommand{\numtargetpoints}{T}

\newcommand{\modelset}{\mathcal{M}}
\newcommand{\nummodels}{M}
\newcommand{\modelindex}{m}

\newcommand{\errorfunction}{\rho}
\newcommand{\deformablemodelforwardfunction}{\phi}
\newcommand{\deformablemodelbackwardfunction}{\psi}

\newcommand{\reward}{r}
\newcommand{\totalregret}{R}

\newcommand{\utility}{u}
\newcommand{\utilityprocessnoise}{v}
\newcommand{\utilityprocessnoisecovar}{\Sigma}
\newcommand{\utilityprocessscale}{\eta}
\newcommand{\transitionnoisefactor}{\sigma_{tr}}
\newcommand{\utilityobs}{\reward}
\newcommand{\utilityobsnoise}{w}
\newcommand{\observationnoisefactor}{\sigma_{obs}}
\newcommand{\modelchosen}{m}
\newcommand{\correlationfactor}{\xi}

\newcommand{\pseudoinverseweight}{W}
\newcommand{\obstacle}{\mathcal{O}}

\title{Bandit-Based Model Selection for Deformable Object Manipulation}
\author{\dale \and Dmitry Berenson}% <-this % stops a space
\institute{University of Michigan, Ann Arbor MI 48109, USA \\ \email{dmcconac@umich.edu, berenson@eecs.umich.edu}}

\begin{document}
\maketitle

\setlength{\textfloatsep}{8.0pt}

\pagestyle{plain}

\begin{abstract}

We present a novel approach to deformable object manipulation that does not rely on highly-accurate modeling. The key contribution of this paper is to formulate the task as a Multi-Armed Bandit problem, with each arm representing a model of the deformable object. To ``pull'' an arm and evaluate its utility, we use the arm's model to generate a velocity command for the gripper(s) holding the object and execute it. As the task proceeds and the object deforms, the utility of each model can change. Our framework estimates these changes and balances exploration of the model set with exploitation of high-utility models. We also propose an approach based on Kalman Filtering for Non-stationary Multi-armed Normal Bandits (KF-MANB) to leverage the coupling between models to learn more from each arm pull. We demonstrate that our method outperforms previous methods on synthetic trials, and performs competitively on several manipulation tasks in simulation.

% \keywords{
%     Deformable Object Manipulation $\cdot$
%     Bandit Problems $\cdot$
%     Online Learning
% }

\end{abstract}

\section{Introduction}

%Manipulation of deformable objects is an area of robotics research that has garnered attention from many diverse fields in recent years. Examples of deformable manipulation range from \rev{domestic} tasks like folding clothes to time and safety critical tasks such as robotic surgery.

One of the primary challenges in manipulating deformable objects is the difficulty of modeling and simulating them.
%Motivated by applications in computer graphics and surgical training, many methods have been developed for simulating string-like objects \cite{Bergou2008,Rungjiratananon2011}, cloth-like objects \cite{Baraff1998,Goldenthal2007}, and tissue-like objects \cite{Sutherland2006,Chentanez2009,Kim2007}. 
The most common simulation methods use Mass-Spring models \cite{Gibson1997,Essahbi2012}, which are generally not accurate for large deformations \cite{Maris2010}, and Finite-Element models \cite{Muller2002,Bathe2006}, which require significant tuning and are very sensitive to the discretization of the object. Approaches like \cite{Schulman2016,Huang2015} bypass this challenge by using offline demonstrations to teach the robot specific manipulation tasks; however, when a new task is attempted a new training set needs to be generated. In our application we are interested in a way to manipulate a deformable object without a high-fidelity model or training set available \textit{a priori}. For instance, imagine a robot encountering a new piece of clothing for a new task. While it may have models for previously-seen clothes or training sets for previous tasks, there is no guarantee that those models or training sets are appropriate for the new task. Also, depending on the state of the clothing different models may be most useful at different times in the manipulation task.

Rather than assuming we have a high-fidelity model of a deformable object interacting with its environment, our approach is to have multiple models available for use, any one of which may be useful at a given time. We do not assume these models are correct, we simply treat the models as having some measurable \textit{utility} to the task. The \textit{utility} of a given model is the expected reduction in task error when using this model to generate robot motion. As the task proceeds, the utility of a given model may change, making other models more suitable for the current part of the task. However, without testing a model's prediction, we do not know its true utility. Testing every model in the set is impractical, as all models would need to be tested at every step, and performing a test changes the state of the object and may drive it into a local minimum. The key question is then which model should be selected for testing at a given time.

The central contribution of this paper is framing the model selection problem as a Multi-Armed Bandit (MAB) problem where the goal is to find the model that has the highest utility for a given task. An arm represents a single model of the deformable object; to ``pull'' an arm is to use the arm's model to generate and execute a velocity command for the robot. The reward received is the reduction in task error after executing the command. In order to determine which model has the highest utility we need to explore the model space, however we also want to exploit the information we have gained by using models that we estimate to have high utility. One of the primary challenges in performing this exploration versus exploitation trade-off is that our models are inherently coupled and non-stationary; performing an action changes the state of the system which can change the utility of every model, as well as the reward of pulling each arm. While there is work that frames robust trajectory selection as a MAB problem~\cite{Koval2015}, we are not aware of any previous work which either 1) frames model selection for deformable objects as a MAB problem; or 2) addresses the coupling between arms for non-stationary MAB problems.

In our experiments, we show how to formulate a MAB problem with coupled arms for Jacobian-based models. We perform our experiments on three synthetic systems, and on three deformable object manipulation tasks in the Bullet~\cite{Coumans2010} simulator. We demonstrate that formulating model selection as a MAB problem is able to successfully perform all three manipulation tasks. We also show that our proposed MAB algorithm outperforms previous MAB methods on synthetic trials, and performs competitively on the manipulation tasks.

\section{Related Work}

\textit{Deformable Object Modeling}: One of the key challenges in manipulating deformable objects is the difficulty inherent in modeling and simulating them. While there has been some progress towards online modeling of deformable objects~\cite{JochenLang2002,Cretu2008a} these methods rely on a time consuming training phase for each object to be modeled. Of particular interest are Jacobian-based models such as~\cite{Berenson2013} and~\cite{Navarro-Alarcon2013}. In these models we assume that there is some function $F : \robotconfigspace \rightarrow \reals^N$ which maps a configuration of $\numgrippers$ robot grippers $\robotconfig \in \robotconfigspace$ to a parameterization of the deformable object $\deformconfig \in \reals^N$, where $N$ is the dimensionality of the parameterization of the deformable object.  These models are then linearized by calculating an approximation of the the Jacobian of $F$:
\begin{align*}
    \deformconfig                               &= F(\robotconfig) \\
    \frac{\partial \deformconfig}{\partial t}   &= \frac{\partial F(\robotconfig)}{\partial \robotconfig} \frac{\partial \robotconfig}{\partial t} \\
    \deformvelocity                             &= J(q) \robotvelocity \enspace .\numberthis
    \label{eqn:jacobian}
\end{align*}

Computation of an exact Jacobian $J(\robotconfig)$ at a given configuration $\robotconfig$ is often computationally intractable and requires high-fidelity models and simulators, so instead approximations are frequently used. A shared characteristic of these approximations is some reliance on tuned parameters. This tuning process can be tedious, and in some cases needs to be done on a per-task basis.

In this paper we consider two types of approximate Jacobian models. The first approximation we use is a \textit{diminishing-rigidity Jacobian}~\cite{Berenson2013} which assumes that points on the deformable object that are near a gripper move ``almost rigidly'' with respect to the gripper while points that are further away move ``less rigidly''. This approximation uses deformability parameters to control how quickly the rigidity decreases with distance. The second approximation we use is an \textit{adaptive Jacobian}~\cite{Navarro-Alarcon2013} which uses online estimation to approximate $J(\robotconfig)$. Adaptive Jacobian models rely on a learning rate to control how quickly the estimation changes from one timestep to the next.

\textit{Model Selection}: In order to accomplish a given manipulation task, we need to determine which type of model to use at the current time to compute the next velocity command, as well as how to set the model parameters. Frequently this selection is done manually, however, there are methods designed to make these determinations automatically. Machine learning techniques such as~\cite{Maron1994,Sparks2015} rely on supervised training data in order to intelligently search for the best regression or classification model, however, it is unclear how to acquire such training data for the task at hand without having already performed the task. The most directly applicable methods come from the Multi-Armed Bandit (MAB) literature~\cite{Auer2002,Gittins2011,Whittle1988}. In this framework there are multiple actions we can take, each of which provides us with some reward according to an unknown probability distribution. The problem then is to determine which action to take (which arm to pull) at each time step in order to maximize reward.

The MAB approach is well-studied for problems where the reward distributions are \textit{stationary}; i.e. the distributions do not change over time~\cite{Auer2002,Agrawal2012}. This is not the case for deformable object manipulation; consider the situation where the object is far away from the goal versus the object being at the goal. In the first case there is a possibility of an action moving the object closer to the goal and thus achieving a positive reward; however, in the second case any motion would, at best, give zero reward.
%In the \textit{contextual bandits} \cite{Agrawal2012confirmthis,Slivkins2009} variation of the MAB problem, additional contextual information or features are observed at each timestep, which can be used to determine which arm to pull. In order to use contextual bandits for a given task, a set of features would need to be engineered, however it is not clear what features to use.

Recent work~\cite{Granmo2010} on non-stationary MAB problems offer promising results that utilize independent Kalman filters as the basis for the estimation of a non-stationary reward distribution for each arm. This algorithm (KF-MANB) provides a Bayesian estimate of the reward distribution at each timestep, assuming that the reward is normally distributed. KF-MANB then performs Thompson sampling~\cite{Agrawal2012} to select which arm to pull, choosing each in proportion to the belief that it is the optimal arm. We build on this approach in this paper to produce a method that also accounts for dependencies between arms by approximating the coupling between arms at each timestep.

For the tasks we address, the reward distributions are both non-stationary as well as \textit{dependent}. Because all arms are operating on the same physical system, pulling one arm both gives us information about the distributions over other arms, as well as changing the future reward distributions of all arms. While work has been done on dependent bandits \cite{Pandey2007,Langford2008}, we are not aware of any work addressing the combination of non-stationary and dependent bandits. Our method for model selection is inspired by KF-MANB, however we directly use coupling between models in order to form a joint reward distribution over all models. This enables a pull of a single arm to provide information about all arms, and thus we spend less time exploring the model space and more time exploiting useful models to perform the manipulation task.

\section{Problem Statement}

Let the robot be represented by a set of $\numgrippers$ grippers with configuration $\robotconfig \in \robotconfigspace$.  We assume that the robot configuration can be measured exactly; in this work we assume the robot to be a set of free floating grippers; in practice we can track the motion of these with inverse kinematics on a real robot. We use the Lie algebra~\cite{Murray1994} of $\se3$ to represent robot gripper velocities. This is the tangent space of $\se3$, denoted as $\tanse3$. The velocity of a single gripper $\gripperindex$ is then $\grippervelocity_\gripperindex = \begin{bmatrix}v_g^T & \omega_g^T\end{bmatrix}^T \in \tanse{3}$ where $v_g$ and $\omega_g$ are the translational and rotational components of the gripper velocity. We define the velocity of the entire robot to be $\robotvelocity = \begin{bmatrix}\robotvelocity_1^T & \dots & \robotvelocity_\numgrippers^T \end{bmatrix}^T \in \robotvelocityspace$. We define the inner product of two gripper velocities $\grippervelocity_1, \robotvelocity_2 \in \tanse3$ to be $\langle \grippervelocity_1, \grippervelocity_2 \rangle = \langle \grippervelocity_1, \grippervelocity_1 \rangle_{c} = v_1^T v_2 + c \omega_1^T \omega_2$, where $c$ is a non-negative scaling factor relating rotational and translational velocities.

The configuration of a deformable object is a set $\deformconfig \subset \reals^3$ of $\numdeformpoints$ points. We assume that we have a method of sensing $\deformconfig$. To measure the norm of a deformable object velocity $\deformvelocity = \begin{bmatrix} \deformvelocity_1^T & \dots & \deformvelocity_\numdeformpoints^T \end{bmatrix}^T \in \deformconfigspace $ we will use a weighted Euclidean norm
\begin{equation}
    \| \deformvelocity \|^2_\pseudoinverseweight = \sum_{i = 1}^\numdeformpoints w_i \deformvelocity_i^T \deformvelocity_i = \deformvelocity^T \diag{(\pseudoinverseweight)} \deformvelocity
\end{equation}
where $W = \begin{bmatrix}w_1 & \dots & w_\numdeformpoints \end{bmatrix}^T \in \reals^\numdeformpoints$ is a set of non-negative weights. The rest of the environment is denoted $\obstacle$ and is assumed to be both static, and known exactly.

Let a \textit{deformation model} be defined as a function $\deformablemodelforwardfunction : \robotvelocityspace \rightarrow \deformconfigspace$ which maps a change in robot configuration $\robotvelocity$ to a change in object configuration $\deformvelocity$. Let $\modelset$ be a set of $\nummodels$ deformable models which satisfy this definition. Each model is associated with a robot command function $\deformablemodelbackwardfunction : \deformconfigspace \times \reals^\numdeformpoints \rightarrow \robotvelocityspace$ which maps a desired deformable object velocity $\deformvelocity$ and weight $\pseudoinverseweight$ (Sec.~\ref{sec:desired_direction}) to a robot velocity command $\robotvelocity$. $\deformablemodelforwardfunction$ and $\deformablemodelbackwardfunction$ also take the object and robot configuration $(\deformconfig,\robotconfig)$ as additional input, however this is omitted for clarity. When a model $\modelindex$ is selected for testing, the model generates a gripper command
\begin{equation}
    \robotvelocity_{\modelindex}(t) = \deformablemodelbackwardfunction_\modelindex(\deformvelocity(t), \pseudoinverseweight(t))
    \label{eqn:robotvelocity}
\end{equation}
which is then executed for one unit of time, moving the deformable object to configuration $\deformconfig(t+1)$.

The problem we address in this paper is which model $\modelindex \in \modelset$ to select in order to to move $\numgrippers$ grippers such that the points in $\deformconfig$ align as closely as possible with some task-defined set of $\numtargetpoints$ target points $\deformtarget \subset \reals^3$, while avoiding gripper collision and excessive stretching of the deformable object. Each task defines a function $\errorfunction$ which measures the alignment error between $\deformconfig$ and $\deformtarget$. The method we present is a local method which picks a single model $\modelindex_{*}$ at each timestep to treat as the true model. This model is then used to reduce error as much as possible while avoiding collision and excessive stretching. 
\begin{equation}
    \modelindex_* = \argmin_{\modelindex \in \modelset} \errorfunction(\deformtarget, \deformconfig(t+1))
    \label{eqn:modelselection}
\end{equation}
We show that this problem can be treated as an instance of the multi-arm non-stationary dependent bandit problem.

\section{Bandit-Based Model Selection}

The primary difficulty with solving~\eqref{eqn:modelselection} directly is that the effectiveness of a particular model in minimizing error is unknown. It may be the case that no model in the set produces the optimal option, however, this does not prevent a model from being useful. In particular the \textit{utility} of a model may change from one task to another, and from one configuration to another as the deformable object changes shape, and moves in and out of contact with the environment. We start by defining the utility $\utility_\modelindex(t) \in \reals$ of a model as the expected improvement in task error $\errorfunction$ if model $\modelindex$ is used to generate a robot command at time $t$. If we know which model has the highest utility then we can solve~\eqref{eqn:modelselection}. This leads to a classic exploration versus exploitation trade-off where we need to explore the space of models in order to learn which one is the most useful, while also exploiting the knowledge we have already gained.  The multi-armed bandit framework is explicitly designed to handle this trade-off.

In the MAB framework, each arm represents a model in~$\modelset$; to pull arm $\modelindex$ is to command the grippers with velocity $\robotvelocity_\modelindex(t)$ (Eq.~\ref{eqn:robotvelocity}) for 1 unit of time. We then define the \textit{reward} $\reward_\modelindex(t+1)$ after taking action $\robotvelocity_\modelindex(t)$ as the improvement in error
\begin{equation}
    \reward_\modelindex(t+1) = \errorfunction(t) - \errorfunction(t+1) = \utility_\modelindex(t) + \utilityobsnoise
    \label{eqn:observedreward}
\end{equation}
where $\utilityobsnoise$ is a zero-mean noise term. The goal is to pick a sequence of arm pulls to minimize total expected regret $\totalregret(T_f)$ over some (possibly infinite) horizon $T_f$
\begin{equation}
    E[\totalregret(T_f)] = \sum_{t=1}^{T_f} (E[\reward^*(t)] - E[\reward(t)])
    \label{eqn:totalregret}
\end{equation}
where $\reward^*(t)$ is the reward of the best model at time $t$. The next section describes how to use bandit-based model selection for deformable object manipulation.

\section{MAB Formulation for Deformable Object Manipulation}

\begin{wrapfigure}[20]{r}{0.6\textwidth}
    \vspace{-0.62in}
    \begin{minipage}{0.6\textwidth}
        \begin{algorithm}[H]
            \caption{MainLoop$(\obstacle, \beta, \lambda)$}
            \begin{algorithmic}[1]
                \State $t \gets 0$
                \State $\relaxeddistancematrix \gets$ GeodesicDistanceMatrix$(\deformconfig_{relaxed})$
                \State $\modelset \gets$ InitializeModels$(\relaxeddistancematrix)$
                \State InitialzeBanditAlgorithm()
                \State $\deformconfig(0) \gets$ SensePoints()
                \State $\robotconfig(0) \gets$ SenseRobotConfig()
                \While{true}
                    \State $\modelchosen \gets $ SelectArmUsingBanditAlgorithm()
                    
                    \State $\deformtarget \gets$ GetTargets()
                    \State $\deformvelocity_e, \pseudoinverseweight_e \gets$ ErrorCorrection$(\deformconfig(t), \deformtarget)$
                    \State $\deformvelocity_s, \pseudoinverseweight_s \gets$ StretchingCorrection$(\relaxeddistancematrix, \lambda, \deformconfig(t))$
                    \State $\deformvelocity_d, \pseudoinverseweight_d \gets$ CombineTerms$(\deformvelocity_e, \pseudoinverseweight_e, \deformvelocity_s, \pseudoinverseweight_s)$
        
                    \State $\robotvelocity_d \gets \deformablemodelbackwardfunction_m(\deformvelocity_d, \pseudoinverseweight_d)$
                    \State $\robotvelocity \gets$ ObstacleRepulsion$(\robotvelocity_d, \obstacle, \beta)$
                    \State CommandConfiguration$(\robotconfig(t) + \robotvelocity)$
        
                    \State $\deformconfig(t + 1) \gets$ SensePoints$()$
                    \State $\robotconfig(t + 1) \gets$ SenseRobotConfig$()$
                    \State UpdateBanditAlgorithm$()$
                    
                    \State $t \gets t + 1$
                \EndWhile
            \end{algorithmic}
            \label{alg:mainloop}
        \end{algorithm}
    \end{minipage}
\end{wrapfigure}
Our algorithm~(Alg.~\ref{alg:mainloop}) can be broken down into four major sections and an initialization block. In the initialization block we pre-compute the geodesic distance between every pair of points in $\deformconfig$ when the deformable object is in its ``natural'' or ``relaxed'' state and store the result in $\relaxeddistancematrix$. These distances are used to construct the deformation models~(Sec.~\ref{sec:jacobian_models}), as well as to avoid overstretching the object~(Sec.~\ref{sec:desired_direction}).
At each iteration we: 
1) pick a model to use to achieve the desired direction~(Sec.~\ref{sec:bandit_algorithms}); 
2) compute the task-defined desired direction to move the deformable object~(Sec.~\ref{sec:desired_direction}); 
3) generate a velocity command using the chosen model~(Sec.~\ref{sec:jacobian_models}); 
4) modify the command to avoid obstacles~(Sec.~\ref{sec:desired_direction});
and 5) update bandit algorithm parameters~(Sec.~\ref{sec:bandit_algorithms}).

\subsection{Algorithms for MAB}
\label{sec:bandit_algorithms}

Previous solutions~\cite{Auer2002,Granmo2010} to minimizing~\eqref{eqn:totalregret} assume that rewards for each arm are normally and independently distributed and then estimate the mean and variance of each Gaussian distribution.  We test three algorithms in our experiments: Upper Confidence Bound for normally distributed bandits UCB1-Normal, Kalman Filter Based Solution to Non-Stationary Multi-arm Normal Bandits (KF-MANB), and our extension of KF-MANB, Kalman Filter Based Solution to Non-Stationary Multi-arm Normal Dependent Bandit (KF-MANDB).

\textit{UCB1-Normal}:
The UCB1-Normal algorithm~\cite{Auer2002} treats each arm (model) as independent, estimating an optimistic Upper Confidence Bound (UCB) for the utility of each model. The model with the highest UCB is used to command the robot at each timestep. This algorithm assumes that the utility of each model is stationary, gradually shifting from exploration to exploitation as more information is gained. While our problem is non-stationary and dependant, we use UCB1-Normal as a baseline algorithm to compare against due to its prevalence in previous work. The algorithm is shown in App.~\ref{apx:ucb1normal} for reference.

\textit{KF-MANB}:
The Kalman Filter Based Solution to Non-Stationary Multi-arm Bandit (KF-MANB) algorithm~\cite{Granmo2010} uses independent Kalman filters to estimate the utility distribution of each model, and then uses Thompson sampling~\cite{Agrawal2012} to chose which model to use at each timestep. Because this algorithm explicitly allows for non-stationary reward distributions, it is able to ``switch'' between models much faster than UCB1-Normal. The KF-MANB algorithm is shown in App.~\ref{apx:ucb1normal} for reference.

\textit{KF-MANDB}:
We also propose a variant of KF-MANB, replacing the independent Kalman filters with a single joint Kalman filter. This enables us to capture the correlations between models, allowing us to learn more from each pull. We start by defining utility as a linear system with Gaussian noise with process model $\utility(t+1) = \utility(t) + \utilityprocessnoise$ and observation model $\utilityobs(t) = C\utility(t) + \utilityobsnoise$ where $\utility(t)$ is our current estimate of the relative utility of each model, while $\utilityprocessnoise$ and $\utilityobsnoise$ are zero-mean Gaussian noise terms. $C$ is a row vector with a 1 in the column of the model we used and zeros elsewhere. The variance on $\utilityobsnoise$ is defined as $\observationnoisefactor^2 \utilityprocessscale^2$. $\utilityprocessscale$ is a tuning parameter to scale the covariance to match the reward scale of the specific task, while $\observationnoisefactor$ controls how much we believe each new observation.

To define the process noise $\utilityprocessnoise$ we want to leverage correlations between models; if two model predictions are similar, the utility of these models is likely correlated. To measure the similarity between two models $i$ and $j$ we use the angle between their gripper velocity commands $\robotvelocity_{i}$ and $\robotvelocity_{j}$.
%$\robotvelocity_{d,i} = \deformablemodelbackwardfunction_i(\deformvelocity_d, \pseudoinverseweight_d)$ and $\robotvelocity_{d,j} = \deformablemodelbackwardfunction_j(\deformvelocity_d, \pseudoinverseweight_d)$.
This similarity is then used to directly construct a covariance matrix for each arm pull:
\begin{equation}
\begin{split}
    \utilityprocessnoise            &\sim \normal{0}{\transitionnoisefactor^2 \utilityprocessscale^2 (\correlationfactor \utilityprocessnoisecovar + \left(1 - \correlationfactor\right) \eye)}\\
    \utilityprocessnoisecovar_{i,j} & = \frac{\langle \robotvelocity_{i}, \robotvelocity_{j} \rangle}{\| \robotvelocity_{i} \| \| \robotvelocity_{j} \|} = \cos \theta_{i,j} \enspace.
\label{eqn:processnoise}
\end{split}
\end{equation}
$\transitionnoisefactor$ is the standard Kalman Filter transition noise factor tuning parameter. $\correlationfactor \in [0,1]$ is the correlation strength factor; larger $\correlationfactor$ gives more weight to the arm correlation, while smaller $\correlationfactor$ gives lower weight. When $\correlationfactor$ is zero then KF-MANDB will have the same update rule as KF-MANB, thus we can view KF-MANDB as a generalizion of KF-MANB, allowing for correlation between arms.

After estimating the utility of each model and the noise parameters at the current timestep, these values are then passed into a Kalman filter which estimates a new joint distribution. The next step is the same as KF-MANB; we draw a sample from the resulting distribution, then use the model that yields the largest sample to generate the next robot command. In this way we automatically switch between exploration and exploitation as the system evolves; if we are uncertain of the utility of our models then we are more likely to choose different models from one timstep to the next. If we believe that we have accurate estimates of utility, then we are more likely to choose the model with the highest utility.

\subsection{Determining $\robotvelocity$}%a Desired Direction $\deformvelocity_d(t)$ and Weight $\pseudoinverseweight_d(t)$}
\label{sec:desired_direction}

\subsubsection{Error Correction}

% \begin{wrapfigure}[21]{r}{0.5\textwidth}
\begin{wrapfigure}[38]{r}{0.5\textwidth}
    \begin{minipage}{0.5\textwidth}
        \centering
        \includegraphics[width=1.8in]{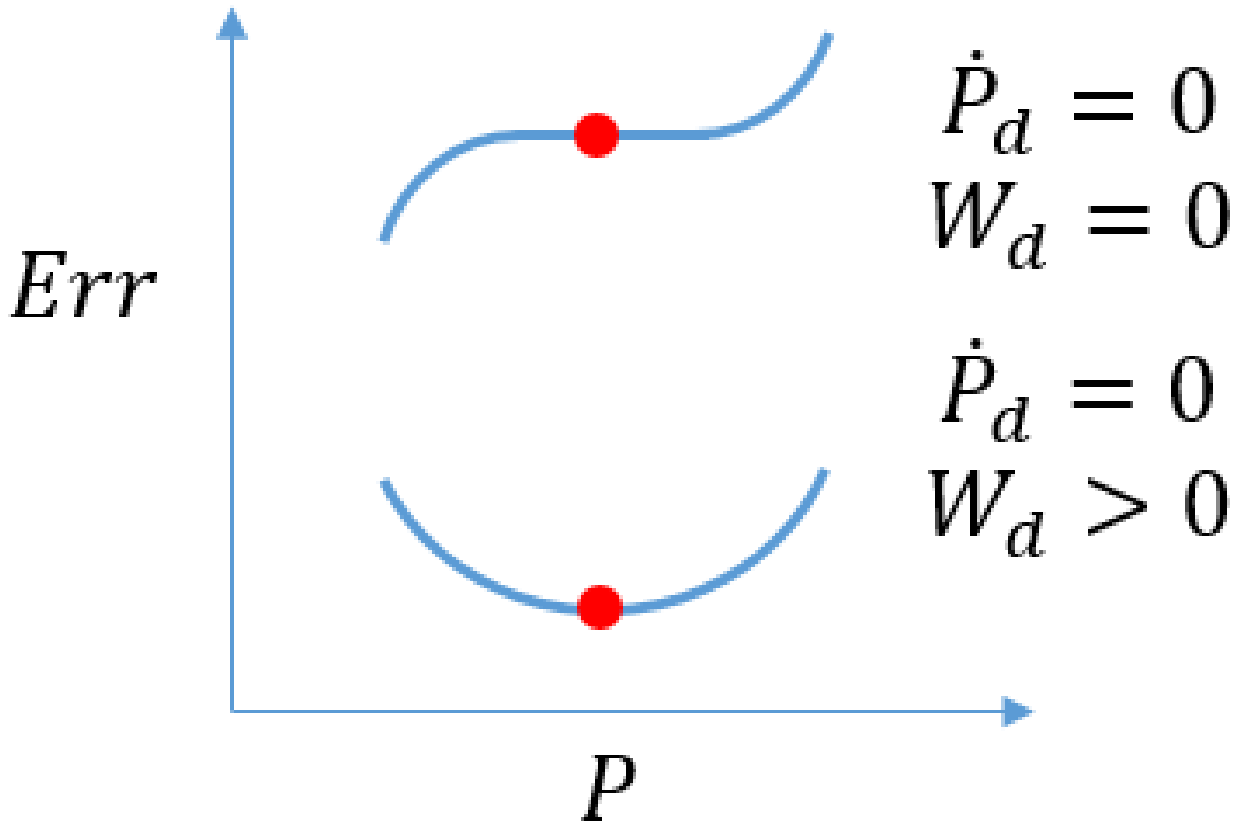}
        \caption{Top Line: moving the point does not change the error, thus the desired movement is zero, however, it is not important to achieve zero movement, thus $W_d = 0$.  Bottom Line: error is at a local minimum; thus moving the point increases error.}
        \vspace{-0.2in}
        \label{fig:error_examples}
    \end{minipage}
    \vspace{-0.55in}
    \begin{minipage}{0.5\textwidth}
        \begin{algorithm}[H]
            \caption{ErrorCorrection$(\deformconfig, \deformtarget)$}
            \begin{algorithmic}[1]
                \State $\deformvelocity_e \gets \boldsymbol 0_{\deformconfigspacesize \times 1}$, $\pseudoinverseweight_e \gets \boldsymbol 0_{\numdeformpoints \times 1}$
                \For{$i \in \{1,2,\dots,\numtargetpoints \}$}
                    \State $k \gets \argmin_{j \in \{ 1,2,\dots,\numdeformpoints \}} \| \deformtarget_i - \deformconfig_j \|$
                    \State $\deformvelocity_{e,k} \gets \deformvelocity_{e,k} + \deformtarget_i - \deformconfig_k$
                    \State $\pseudoinverseweight_{e,k} \gets \max (\pseudoinverseweight_{e,k}, \| \deformtarget_i - \deformconfig_k \|)$
                \EndFor
                \State \Return $\{ \dot \deformconfig_e, \pseudoinverseweight_e \}$
            \end{algorithmic}
            \label{alg:error_correction}
        \end{algorithm}
        \vspace{-0.57in}
        \begin{algorithm}[H]
            \caption{StretchingCorrection$(\relaxeddistancematrix, \lambda, \deformconfig)$}
            \begin{algorithmic}[1]
                \State $E \gets$ EuclidianDistanceMatrix$(\deformconfig)$
                \State $\deformvelocity_s \gets \boldsymbol 0_{3\numdeformpoints \times 1}$, $\pseudoinverseweight_s \gets \boldsymbol 0_{\numdeformpoints \times 1}$
                \State $\Delta \gets E - \relaxeddistancematrix$
                \For{$i \in \{1,2,\dots,\numdeformpoints \}$}
                    \For{$j \in \{i+1,\dots,\numdeformpoints \}$}
                        \If{$\Delta_{i,j} > \lambda$}
                            \State $v \gets \Delta_{i,j}(\deformconfig_j - \deformconfig_i)$
                            \State $\deformvelocity_{s,i} \gets \deformvelocity_{s,i} + \frac{1}{2}v$
                            \State $\deformvelocity_{s,j} \gets \deformvelocity_{s,j} - \frac{1}{2}v$
                            \State $\pseudoinverseweight_{s,i} \gets \max (\pseudoinverseweight_{s,i}, \Delta_{i,j})$
                            \State $\pseudoinverseweight_{s,j} \gets \max (\pseudoinverseweight_{s,j}, \Delta_{i,j})$
                        \EndIf
                    \EndFor
                \EndFor
                \State \Return $\{ \deformvelocity_s, \pseudoinverseweight_s \}$
            \end{algorithmic}
            \label{alg:stretching_correction}
        \end{algorithm}
    \end{minipage}
    \vspace{-0.62in}
\end{wrapfigure}
We build on previous work~\cite{Berenson2013}, splitting the desired deformable object movement into two parts: an error correction part and a stretching correction part. When defining the direction we want to move the deformable object to minimize error we calculate two values; which direction to move the deformable object points $\deformvelocity_e$ and the importance of moving each deformable object point $\pseudoinverseweight_e$. This is analogous to computing the gradient of error, as well as an ``importance factor'' for each part of the gradient. We need these weights to be able to differentiate between points of the object where the error function is a plateau versus points where the error function is at a local minimum~(Fig.~\ref{fig:error_examples}). Typically this is achieved using a Hessian, however our error function does not have a second derivative at many points. We use the \texttt{ErrorCorrection} (Alg.~\ref{alg:error_correction}) function to calculate these values. Each target point $\deformtarget_i \in \deformtarget$ defines a potential field, pulling the nearest point on the deformable object $\deformconfig_k$ towards $\deformtarget_i$. $\pseudoinverseweight_e$ is set to the maximum distance $\deformconfig_k$ is being pulled by any target point. This allows $\pseudoinverseweight_e$ to be insensitive to changes in discretization.

\subsubsection{Stretching Correction}

Our algorithm for stretching correction is similar to that found in~\cite{Berenson2013}, with the addition of a weighting term $\pseudoinverseweight_s$, and a change in how we combine the two terms. We use the \texttt{StretchingCorrection} function (Alg.~\ref{alg:stretching_correction}) to compute $\deformvelocity_s$ and $\pseudoinverseweight_s$ based on a task-defined stretching threshold $\lambda \geq 0$. First we compute the distance between every two points on the object and store the result in $E$. We then compare $E$ to $D$ which contains the relaxed lengths between every pair of points. If any two points are stretched by more than $\lambda$, we attempt to move the points closer to each other. We use the same strategy for setting the importance of this stretching correction $\pseudoinverseweight_s$ as we use for error correction. When combining stretching correction and error correction terms (Alg.~\ref{alg:combine_terms}) we prioritize stretching correction, accepting only the portion of the error correction that is orthogonal to the stretching correction term for each point.

\subsubsection{Obstacle Avoidance}

\begin{wrapfigure}{r}{0.5\textwidth}
    \vspace{-0.55in}
    \begin{minipage}{0.5\textwidth}
        \vspace{-0.05in}
        \begin{algorithm}[H]
            \caption{CombineTerms$(\deformvelocity_e, \pseudoinverseweight_e, \deformvelocity_s, \pseudoinverseweight_s)$}
            \begin{algorithmic}[1]
                \For{$i \in \{ 1,2,\dots,\numdeformpoints \}$}
                    \State $\deformvelocity_{d,i} \gets \deformvelocity_{s,i} + \left( \deformvelocity_{e,i} - \proj_{\deformvelocity_{s,i}} \deformvelocity_{e,i} \right)$
                    \State $\pseudoinverseweight_{d,i} \gets \pseudoinverseweight_{s,i} + \pseudoinverseweight_{e,i}$
                \EndFor
                \State \Return $\{ \deformvelocity_d, \pseudoinverseweight_d \}$
            \end{algorithmic}
            \label{alg:combine_terms}
        \end{algorithm}
        \vspace{-0.57in}
        \begin{algorithm}[H]
            \caption{ObstacleRepulsion$(\obstacle, \beta)$}
            \begin{algorithmic}[1]
                \For{$\gripperindex \in \{1,2,\dots, \numgrippers\}$}
                    \State $J_{p^g}, \dot x _{p^g}, d_g \gets$ Proximity$(\obstacle, \gripperindex)$
                    \State $\gamma \gets e^{-\beta d_g}$
                    \State $\robotvelocity_{c,g} \gets J_{p^g}^+ \dot x_{p^g}$
                    \State $\robotvelocity_{c,g} \gets \frac{\maxgrippervelobstacle}{\| \robotvelocity_{c,g} \|} \robotvelocity_{c,g}$
                    \State  $\robotvelocity_{\gripperindex} \gets \gamma \left( \robotvelocity_{c,g} + \left( \eye - J_{p^g}^+ J_{p^g} \right) \robotvelocity_\gripperindex \right) + (1-\gamma)\robotvelocity_\gripperindex$
                \EndFor        
                \State \Return $\robotvelocity$
            \end{algorithmic}
            \label{alg:obstaclerepulsion}
        \end{algorithm}
        \vspace{-0.8in}
    \end{minipage}
\end{wrapfigure}
In order to guarantee that the grippers do not collide with any obstacles, we use the same strategy from~\cite{Berenson2013}, smoothly switching between collision avoidance and other objectives (see Alg.~\ref{alg:obstaclerepulsion}). For every gripper $\gripperindex$ and an obstacle set $\obstacle$ we find the distance $d_\gripperindex$ to the nearest obstacle, a unit vector $\dot x_{p_\gripperindex}$ pointing from the obstacle to the nearest point on the gripper, and a Jacobian $J_{p^\gripperindex}$ between the gripper's DOF and the point on the gripper. The \texttt{Proximity} function is shown in Appendix~\ref{apx:obstacle_proximity}. $\beta > 0$ sets the rate at which we change between servoing and collision avoidance objectives. $\maxgrippervelobstacle > 0$ is an internal parameter that sets how quickly we move the robot away from obstacles.

\subsection{Jacobian Models}
\label{sec:jacobian_models}

Every model must define a prediction function $\deformablemodelforwardfunction(\robotvelocity)$ and has an associated robot command function $\deformablemodelbackwardfunction(\deformvelocity, \pseudoinverseweight)$. This paper focuses on Jacobian-based models whose basic formulation Eq.~\eqref{eqn:jacobian} directly defines the deformation model $\deformablemodelforwardfunction$
\begin{equation}
    \deformablemodelforwardfunction(\robotvelocity) = J \robotvelocity.\enspace
\end{equation}
When defining the robot command function $\deformablemodelbackwardfunction$, we use the weights $\pseudoinverseweight$ to focus the robot motion on the important part of $\deformvelocity$. This is done by using a weighted norm in a standard minimization problem
% \begin{equation}
% \begin{aligned}
%     \deformablemodelbackwardfunction(\deformvelocity, \pseudoinverseweight) = 
%         &\argmin_{\robotvelocity }   & & \| J \robotvelocity - \deformvelocity \|^2_{\pseudoinverseweight} \\
%         & \text{subject to}          & & \| \robotvelocity \|^2 < \maxgrippervelservo^2
% \end{aligned}
% \end{equation}
\begin{equation}
    \deformablemodelbackwardfunction(\deformvelocity, \pseudoinverseweight) = \argmin_{\robotvelocity } \| J \robotvelocity - \deformvelocity \|^2_{\pseudoinverseweight} \mbox{ s.t. } \| \robotvelocity \|^2 < \maxgrippervelservo^2. \enspace
    \label{eqn:jacobianbackwardfunction}
\end{equation}
We also need to ensure that the grippers do not move too quickly, so we add the constraint that the robot moves no more than $\maxgrippervelservo > 0$. To solve \eqref{eqn:jacobianbackwardfunction} we use the Gurobi~\cite{Gurobi2016} optimizer. We use two different Jacobian approximation methods in our model set; a diminishing rigidity Jacobian, and an adaptive Jacobian, which are described below.

\subsubsection{Diminishing Rigidity Jacobian}

The key assumption used by this method~\cite{Berenson2013} is \textit{diminishing rigidity}: the closer a gripper is to a particular part of the deformable object, the more that part of the object moves in the same way that the gripper does (i.e. more ``rigidly''). The further away a given point on the object is, the less rigidly it behaves; the less it moves when the gripper moves. Details of how to construct a diminishing rigidity Jacobian are in Appendix~\ref{apx:diminishing_rigidity}. This approximation depends on two parameters $k_{trans}$ and $k_{rot}$ which control how the translational and rotational rigidity scales with distance. Small values entail very rigid objects; high values entail very deformable objects.

\subsubsection{Adaptive Jacobian}

A different approach is taken in~\cite{Navarro-Alarcon2013}, instead using online estimation to approximate $J(q)$.
In this formulation we start with some estimate of the Jacobian $\tilde J(0)$ at time $t = 0$ and then use the Broyden update rule~\cite{Broyden1965} to update $\tilde J(t)$ at each timestep $t$
\begin{equation}
    \tilde J(t) = \tilde J(t-1) + \Gamma \frac{\left( \deformvelocity(t) - \tilde J(t-1) \robotvelocity(t) \right)}{\robotvelocity(t)^T \robotvelocity(t)} \robotvelocity(t)^T \enspace.
\end{equation}
This update rule depends on a update rate $\Gamma \in (0, 1]$ which controls how quickly the estimate shifts between timesteps.

\section{Experiments and Results}

We test our method on three synthetic tests and three deformable object manipulation tasks in simulation. The synthetic tasks show that the principles we use to estimate the coupling between models are reasonable; while the simulated tasks show that our method is effective at performing deformable object manipulation tasks.

\subsection{Synthetic Tests}

For the synthetic tests, we set up an underactuated system that is representative of manipulating a deformable object with configuration $y \in \reals^n$ and control input $\dot x \in \reals^m$ such that $m < n$ and $\dot y = J \dot x$. To construct the Jacobian of this system we start with $J = \begin{bmatrix}\eye_{m \times m} \\ \mathbf{0}_{(n-m) \times m} \end{bmatrix}$ and add uniform noise drawn from $[-0.1, 0.1]$ to each element of $J$. The system configuration starts at $\begin{bmatrix}10 & \dots & 10\end{bmatrix}^T$ with the target configuration set to the origin. Error is defined as $\errorfunction(t) = \| y(t) \|$, and the desired direction to move the system at each timestep is $\dot y_d(t) = - y(t)$. These tasks have no obstacles or stretching, thus $\beta, \lambda,$ and $\maxgrippervelobstacle$ are unused. Rather than setting the utility noise scale $\utilityprocessscale$ \textit{a priori}, we use an annealing filter
\begin{equation}
    \utilityprocessscale(t+1) = \max(10^{-10}, 0.9 \utilityprocessscale(t) + 0.1 |\reward(t+1)|) \enspace.
    \label{eqn:jacobian_minimization}
\end{equation}
This enables us to track the changing available reward as the system gets closer to the target. All other parameters are shown in~App~\ref{apx:param_table}.

To generate a model for the model set we start with the true Jacobian $J$ and add uniform noise drawn from $[-0.025, 0.025]$ to each element of $J$. For an individual trial, each bandit algorithm uses the same $J$ and the same model set. Each bandit algorithm receives the same random number stream during a trial, ensuring that a more favourable stream doesn't bias results. We ran one small test using a $3 \times 2$ Jacobian with 10 arms in order to yield results that are easily visualised. The second and third tests are representative of the scale of the simulation experiments, using the same number of models and similar sizes of Jacobian as are used in simulation. A single trial consists of 1000 pulls (1000 commanded actions); each test was performed 100 times to generate statistically significant results. Our results in Table~\ref{tab:synthetic_results} show that KF-MANDB clearly performs the best for all three tests.

\begin{table}[t]
\centering
\caption{Synthetic trial results showing total regret with standard deviation in brackets for all bandit algorithms for 100 runs of each setup.}
\label{tab:synthetic_results}
\begin{tabular}{cccccc}
\hline\noalign{\smallskip}
\# of Models & $n$     & $m$     & UCB1-Normal & KF-MANB     & KF-MANDB \\
\hline\noalign{\smallskip}
10           & 3       & 2       & 4.41 [1.65] & 3.62 [1.73] & 2.99 [1.40] \\
60           & 147     & 6       & 5.57 [1.37] & 4.89 [1.32] & 4.53 [1.42] \\
60           & 6075    & 12      & 4.21 [0.64] & 3.30 [0.56] & 2.56 [0.54] \\
\hline\noalign{\smallskip}
\end{tabular}
\end{table}

\subsection{Simulation Trials}

\begin{figure}[t]
    \centering
    \includegraphics[width=\textwidth]{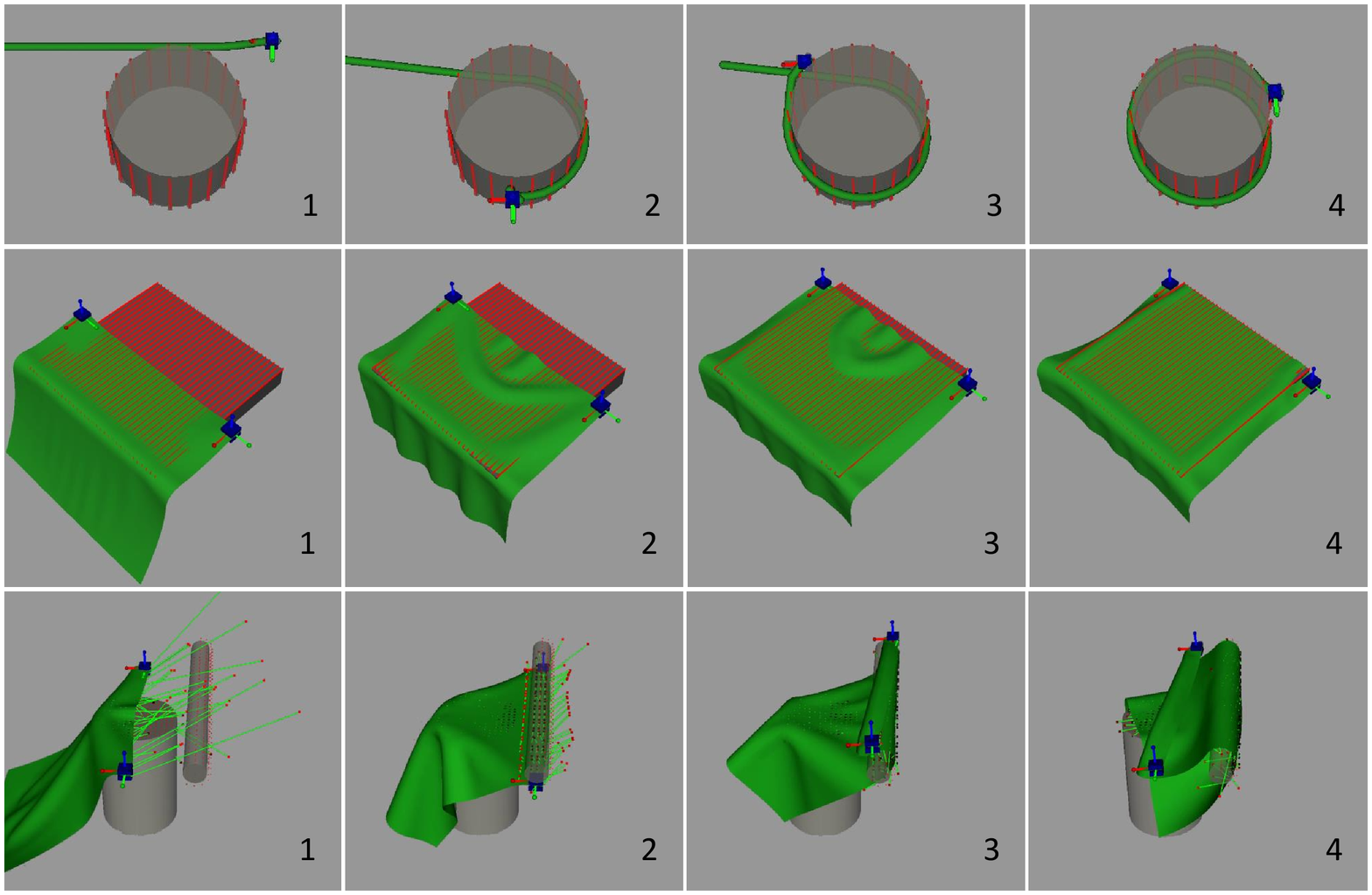}
    \caption{Sequence of snapshots showing the execution of the simulated experiments using the KF-MANDB algorithm. The rope and cloth are shown in green, the grippers is shown in blue, and the target points are shown in red. The bottom row additionally shows $\deformvelocity_d$ as green rays with red tips.}
    \label{fig:simulation_task_screenshots}
\end{figure}

We now demonstrate the effectiveness of multi-arm bandit techniques on three example tasks, show how to encode those tasks for use in our framework, and discuss experimental results. The first task shows how our method can be applied to a rope, with the goal of winding the rope around a cylinder in the environment. The second and third tasks show the method applied to cloth. In the second task, two grippers manipulate the cloth so that it covers a table. In the third task, we perform a two-stage coverage task, covering portions of two different cylinders. In all three tasks, the alignment error $\errorfunction(\deformconfig, \deformtarget)$ is measured as the sum of the distances between every point in $\deformtarget$ and the closest point in $\deformconfig$ in meters. Figure~\ref{fig:simulation_task_screenshots} shows the target points in red, and the deformable object in green. The video accompanying this paper shows the task executions.

All experiments were conducted in the open-source Bullet simulator~\cite{Coumans2010}, with additional wrapper code developed at UC Berkeley. The rope is modeled as a series of 49 small capsules linked together by springs and is 1.225m long. The cloth is modeled as a triangle mesh of size $0.5\text{m} \times 0.5\text{m}$ for the table coverage task, and size $0.5\text{m} \times 0.625\text{m}$ for the two-stage coverage task. We emphasize that our method does not have access to the model of the deformable object or the simulation parameters. The simulator is used as a ``black box'' for testing.

We use models generated using the same parameters for all three tasks with a total of 60 models: 49 diminishing rigidity models with rotation and translational deformability values $k_{trans}$ and $k_{rot}$ ranging from 0 to 24 in steps of 4, as well as 11 adaptive Jacobian models with learning rates $\Gamma$ ranging from $1$ to $10^{-10}$ in multiples of 10. All adaptive Jacobian models are initialized with the same starting values; we use the diminishing rigidity Jacobian for this seed with $k_{trans}=k_{rot}=10$ for the rope experiment and $k_{trans}=k_{rot}=14$ for the cloth experiments to match the best model found in~\cite{Berenson2013}. We use the same strategy for setting $\utilityprocessscale$ as we use for the synthetic tests. App~\ref{apx:param_table} shows all other parameters.

We evaluate results for the MAB algorithms as well as using each of the models in the set for the entire task. To calculate regret for each MAB algorithm, we create copies of the simulator at every timestep and simulate the gripper command, then measure the resulting reward $\reward_\modelindex(t)$ for each model. The reward of the best model $\reward^*(t)$ is then the maximum of individual rewards. As KF-MANB and KF-MANDB are not deterministic algorithms, each task is performed 10 times for these methods. All tests are run on an Intel Xeon E5-2683 v4 processor with 64 GB of RAM. UCB1-Normal and KF-MANB solve Eq.~\eqref{eqn:jacobianbackwardfunction} once per timestep, while KF-MANDB solves it for every model in $\modelset$. Computation times for each test are shown in their respective sections.

\begin{figure}[t]
    \centering
    \vspace{-0.1in}
    \subfloat{
        \includegraphics[width=0.45\textwidth]{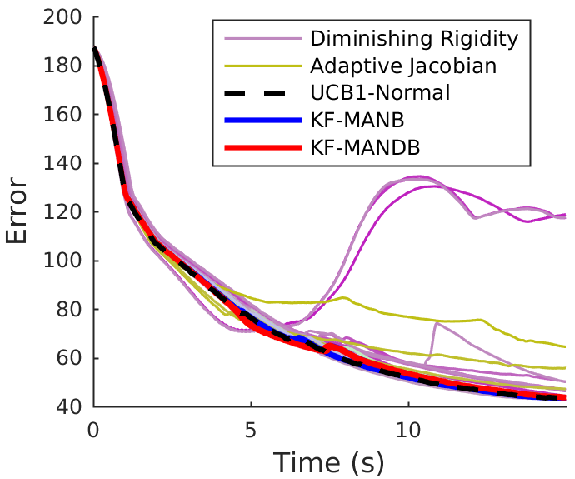}
    }
    \subfloat{
        \includegraphics[width=0.45\textwidth]{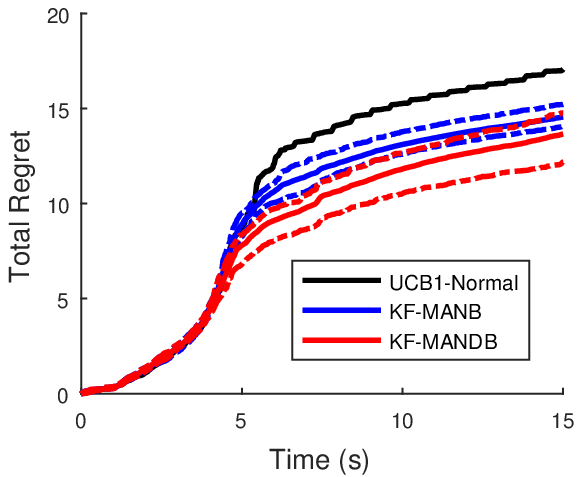}
    }
    \\
    \vspace{-0.15in}
    \subfloat{
        \hspace{-0.1in}
        \includegraphics[width=0.35\textwidth]{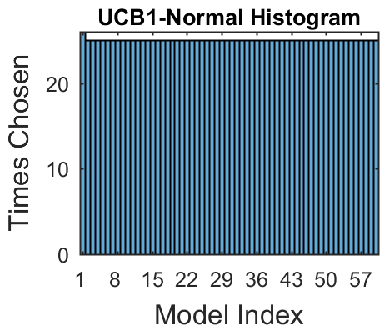}
        \hspace{-0.25in}
    }
    \subfloat{
        \includegraphics[width=0.35\textwidth]{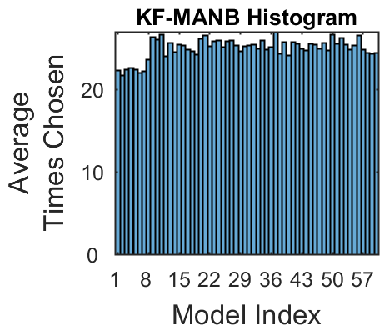}
    }
    \subfloat{
        \hspace{-0.25in}
        \includegraphics[width=0.35\textwidth]{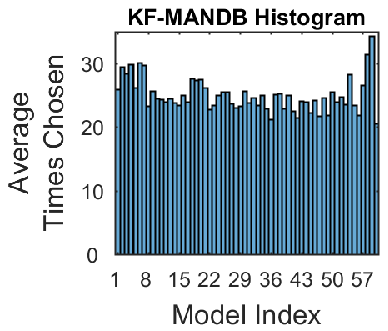}
        \hspace{-0.1in}
    }
    \vspace{-0.1in}
    \caption{Experimental results for the rope-winding task. Top left: alignment error for 10 trials for each MAB algorithm, and each model in the model set when used in isolation. UCB1-Normal, KF-MANB, KF-MANDB lines overlap in the figure for all trials. Top right: Total regret averaged across 10 trials for each MAB algorithm with the minimum and maximum drawn in dashed lines. Bottom row: histograms of the number of times each model was selected by each MAB algorithm; UCB1-Normal (bl), KF-MANB (bm), KF-MANDB (br).}
    \label{fig:ropecylinder_results}
\end{figure}

\textit{Winding a Rope Around a Cylinder}: In the first example task, a single gripper holds a rope that is lying on a table. The task is to wind the rope around a cylinder which is also on the table (see Fig.~\ref{fig:simulation_task_screenshots}). Our results~(Fig.~\ref{fig:ropecylinder_results}) show that at the start of the task all the individual models perform nearly identically, starting to split at 2 seconds (when the gripper first approaches the cylinder) and again at 6 seconds. Despite our model set containing models that are unable to perform the task, our formulation is able to successfully perform the task using all three bandit algorithms. Interestingly, while KF-MANDB outperforms UCB1-Normal and KF-MANB in terms of regret, all three algorithms produce very similar results. Solving Eq.~\eqref{eqn:jacobianbackwardfunction} at each iteration requires an average of 17.3~ms (std. dev. 5.5~ms) for a single model, and 239.5~ms (std. dev. 153.7~ms) for 60 models.

\begin{figure}[t]
    \centering
    \vspace{-0.1in}
    \subfloat{
        \includegraphics[width=0.45\textwidth]{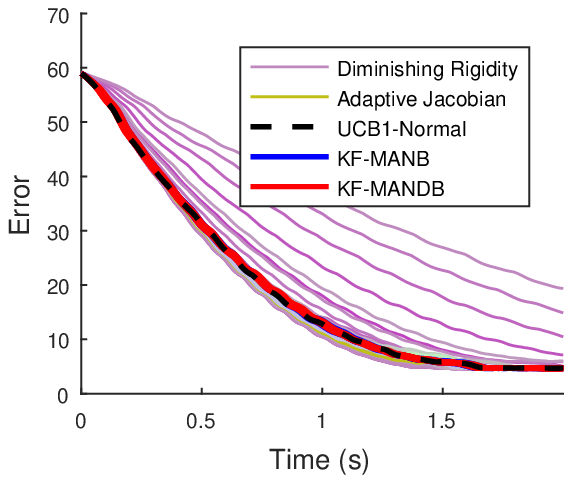}
    }
    \subfloat{
        \includegraphics[width=0.45\textwidth]{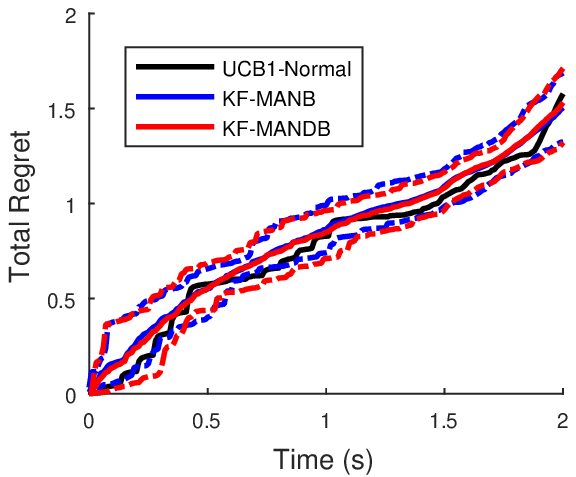}
    }
    \\
    \vspace{-0.15in}
    \subfloat{
        \hspace{-0.1in}
        \includegraphics[width=0.35\textwidth]{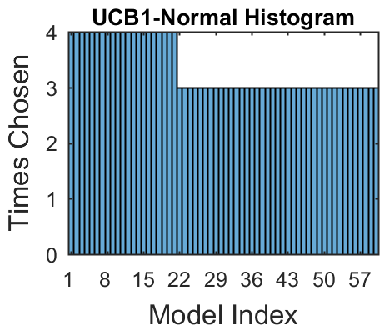}
        \hspace{-0.25in}
    }
    \subfloat{
        \includegraphics[width=0.35\textwidth]{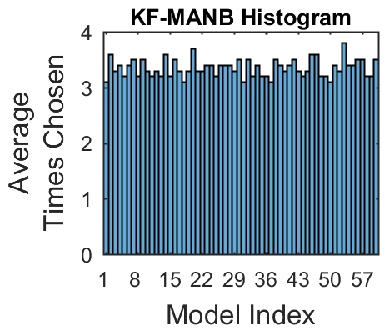}
    }
    \subfloat{
        \hspace{-0.25in}
        \includegraphics[width=0.35\textwidth]{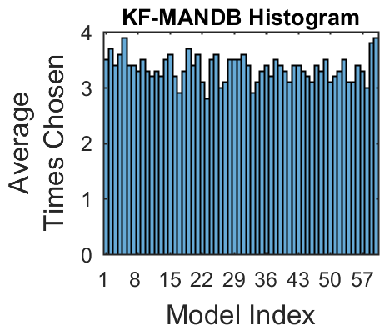}
        \hspace{-0.1in}
    }
    \vspace{-0.1in}
    \caption{Experimental results for the table coverage task. See Fig.~\ref{fig:ropecylinder_results} for description.}
    \label{fig:clothtable_results}
\end{figure}

\textit{Spreading a Cloth Across a Table}: The second scenario we consider is spreading a cloth across a table. In this scenario two grippers hold the rectangular cloth at two corners and the task is to cover the top of the table with the cloth. All of the models are able to perform the task (see Fig.~\ref{fig:clothtable_results}), however, many single-model runs are slower than the bandit methods at completing the task, showing the advantage of the bandit methods. When comparing between the bandit methods, both error and total regret indicate no performance difference between the methods. Solving Eq.~\eqref{eqn:jacobianbackwardfunction} at each iteration requires an average of 89.5~ms (std. dev. 82.4~ms) for a single model, and 605.1~ms (std. dev. 514.3~ms) for 60 models.

\begin{figure}[t]
    \centering
    \vspace{-0.1in}
    \subfloat{
        \includegraphics[width=0.45\textwidth]{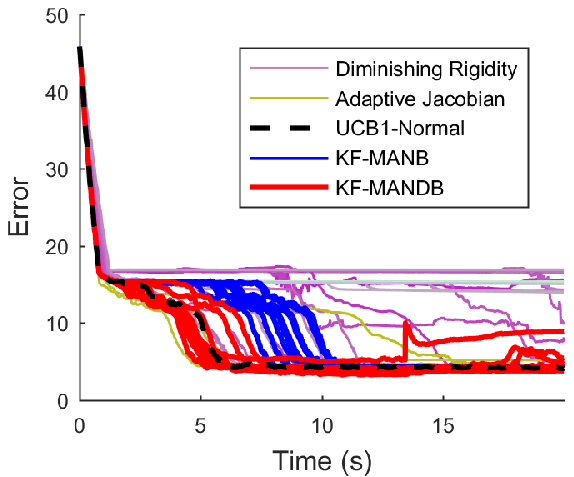}
    }
    \subfloat{
        \includegraphics[width=0.45\textwidth]{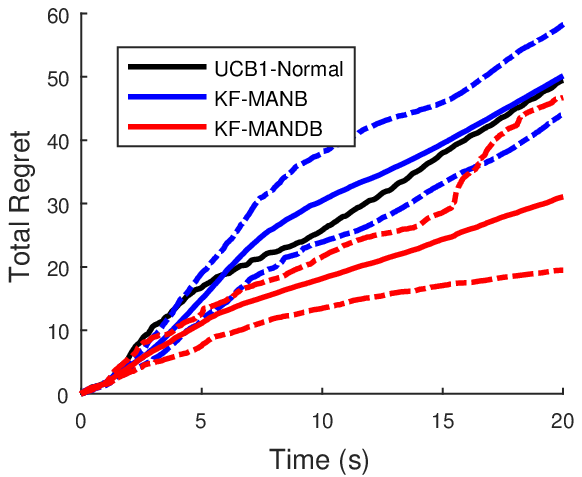}
    }
    \\
    \vspace{-0.15in}
    \subfloat{
        \hspace{-0.1in}
        \includegraphics[width=0.35\textwidth]{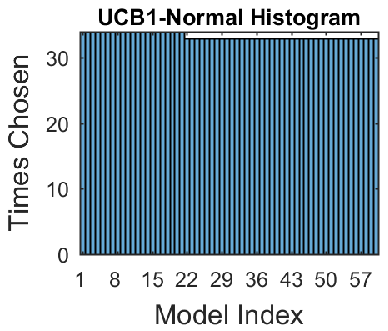}
        \hspace{-0.25in}
    }
    \subfloat{
        \includegraphics[width=0.35\textwidth]{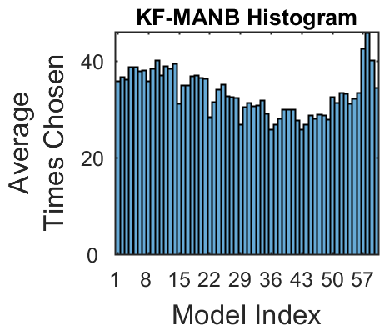}
    }
    \subfloat{
        \hspace{-0.25in}
        \includegraphics[width=0.35\textwidth]{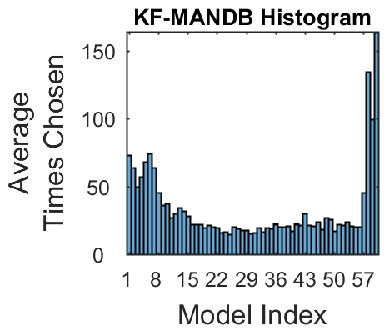}
        \hspace{-0.1in}
    }
    \vspace{-0.1in}
    \caption{Experimental results for the two-part coverage task. See Fig.~\ref{fig:ropecylinder_results} for description.}
    \label{fig:clothwafr_results}
\end{figure}

\textit{Two-Part Coverage Task}: In this experiment, we consider a two-part task. The first part of the task is to cover the top of a cylinder similar to our second scenario. The second part of the task is to cover the far side of a second cylinder. For this task the \texttt{GetTargets} function used previously pulls the cloth directly into the second cylinder. The collision avoidance term then negates any motion in that direction causing the grippers to stop moving. To deal with this, we discretize the free space using a voxel grid, and then use Dijkstra's algorithm to find a collision free path between each cover point and every point in free space. We use the result from Dijkstra's algorithm to define a vector field that pulls the nearest (as defined by Dijkstra's) deformable object point $p_k$ along the shortest collision free path to the target point. This task is the most complex of the three (see Fig.~\ref{fig:clothwafr_results}); many models are unable to perform the task at all, becoming stuck early in the task. We also observe that both KF-MANB and KF-MANDB show a preference for some models over others. Two interesting trials using KF-MANDB stand out; in the first the grippers end up on opposite sides of the second cylinder, in this configuration the physics engine has difficulty resolving the scene and allows the cloth to be pulled straight through the second cylinder. In the other trial the cloth is pulled off of the first cylinder, however KF-MANDB is able to recover, moving the cloth back onto the first cylinder. KF-MANDB and UCB1-Normal are able to perform the task significantly faster than KF-MANB, though all MAB methods complete the task using our formulation. Solving Eq.~\eqref{eqn:jacobianbackwardfunction} at each iteration requires an average of 102.6~ms (std. dev. 30.6~ms) for a single model, and 565.5~ms (std. dev. 389.8~ms) for 60 models.

\vspace{-0.03in}
\section{Conclusion}
\vspace{-0.03in}

We have formulated model selection for deformable object manipulation as a MAB problem. Our formulation enables the application of existing MAB algorithms to deformable object manipulation as well as introduces a novel \textit{utility} metric to measure how useful a model is at performing a given task. We have also presented Kalman Filtering for Non-stationary Multi-arm Normal Dependent Bandits (KF-MANDB) to leverage coupling between dependent bandits to learn more from each arm pull. Our experiments show how to perform several interesting tasks for rope and cloth using our method.

One notable result we observe is that finding and exploiting the best model is less important than avoiding poor models for extended periods of time; in all of the experiments UCB1-Normal never leaves its initial exploration phase, however it is able to successfully perform each task. We believe this is due to many models being able to provide commands that have a positive dot-product with the correct direction of motion.

%Despite this lack of exploitation, UCB1-Normal remains competitive with the other algorithms.

One limitation of KF-MANDB is handling bifurcations; when very small differences in command sent to the robot cause large differences in the result the assumption of coupling between models in KF-MANDB does not hold. In future work we seek to explore how to overcome this limitation, as well as using the predictive accuracy of each model as an additional measure of model coupling. 

\vspace{-0.03in}
\section{Acknowledgements}
\vspace{-0.03in}

This work was supported in part by NSF grants IIS-1656101 and IIS-1551219. We gratefully acknowledge Calder Phillips-Grafflin for his assistance with Bullet.

\bibliographystyle{splncs}
\bibliography{references,all}

\newpage

\appendix

\section{MAB Algorithm Blocks}

\subsection{UCB1-Normal}
\label{apx:ucb1normal}

Reproduced from~\cite{Auer2002}.
\\
\\
\textbf{Loop}: For each $n = 1, 2, \dots$

\begin{itemize}
    \item If there is a machine which has been played less than $\ceil*{8 \log n}$ times then play this machine. If multiple machines qualify, we play the machine that has been played less, selecting the machine with the lower index in the case of a tie.
    \item Otherwise play machine $j$ that maximizes
        \begin{equation}
            \bar x_j + \sqrt{16 \cdot \frac{q_j -n_j \bar x_j^2}{n_j - 1} \cdot \frac{\ln(n-1)}{n_j}}
        \end{equation}
        where $\bar x_j$ is the average reward obtained from machine $j$, $q_j$ is the sum of squared rewards obtained from machine $j$, and $n_j$ is the number of times machine $j$ has been played so far.
    \item Update $\bar x_j$ and $q_j$ with the obtained reward $x_j$.
\end{itemize}

\subsection{KF-MANB}
\label{apx:kfmanb}

\renewcommand{\algorithmicrequire}{\textbf{Input:}}
\renewcommand{\algorithmicensure}{\textbf{Initialization:}}

\begin{minipage}{\textwidth}
\vspace{-0.2in}
\begin{algorithm}[H]
    \caption{KF-MANB - reproduced from~\cite{Granmo2010}}
    \label{alg:kf-manb}
    \begin{algorithmic}
        \Require Number of bandit arms $L$; Observation noise $\sigma_{ob}^2$; Transition noise $\sigma_{tr}^2$.
        \Ensure $\mu_q[1] = \mu_2[1] = \dots = \mu_L[1] = A$; $\sigma_1[1] = \sigma_2[1] = \dots = \sigma_L[1] = B$; \textit{\# Typically, $A$ can be set to $0$, with $B$ being sufficiently large}
        \For{$N = 1,2,\dots$}
            \State{
            \begin{enumerate}
                \item{For each arm $j \in \{1,\dots,L\}$}, draw a value $x_j$ randomly from the associated \textit{normal} distribution $f(x_j;\mu_j[N],\sigma_j[N])$ with the parameters $(\mu_j[N],\sigma_j[N])$.
                \item{Pull the arm $i$ whose drawn $x_i$ is the largest one:
                        \begin{equation*}
                            i = \argmax_{j \in \{1,\dots,L\}} x_j.
                        \end{equation*}
                    }
                \item{Receive reward $\tilde r_i$ from pulling arm $i$, and update parameters as follows:
                        \begin{itemize}
                            \item{Arm $i$:
                                \begin{align*}
                                    \mu_i[N+1]      &= \frac{(\sigma_i^2[N] + \sigma_{tr}^2) \cdot \tilde r_i + \sigma_{ob}^2 \cdot \mu_i[N]}{\sigma_i^2[N] + \sigma_{tr}^2 + \sigma_{ob}^2} \\
                                    \sigma_i^2[N+1] &= \frac{(\sigma_i^2[N] + \sigma_{tr}^2) \sigma_{ob}^2}{\sigma_i^2[N] + \sigma_{tr}^2 + \sigma_{ob}^2}
                                \end{align*}
                            }
                            \item{Arm $j \neq i$:
                                \begin{align*}
                                    \mu_j[N+1]      &= \mu_j[N] \\
                                    \sigma_j^2[N+1] &= \sigma_j[N] + \sigma_{tr}^2
                                \end{align*}
                            }
                        \end{itemize}
                    }
            \end{enumerate}}
        \EndFor
    \end{algorithmic}
\end{algorithm}
\vspace{-0.3in}
\end{minipage}

\section{Diminishing Rigidity Jacobian Construction}
\label{apx:diminishing_rigidity}

For every point $p_i \in \deformconfig$ and every gripper $\gripperindex$ we construct a Jacobian $J_{rigid}(q,i,g)$ such that if $p_i$ was rigidly attached to the gripper $g$ then
\begin{equation}
    \dot p_i = J_{\mathit{rigid}}(q,i,g) \dot \robotconfig_{\gripperindex} = 
    \begin{bmatrix}J_{trans}(q,i,g) & & & J_{rot}(q,i,g)\end{bmatrix}
    \dot \robotconfig_{\gripperindex} \enspace .
\end{equation}
Let $D_{i,g}$ be a measure of the distance between gripper $g$ and point $p_i$. Then the translational rigidity of point $p_i$ with respect to gripper $g$ is defined as
\begin{equation}
    w_{trans}(i,g) = e^{-k_{trans}D_{i,g}}
\end{equation}
and the rotational rigidity is defined as
\begin{equation}
    w_{rot}(i,g) = e^{-k_{rot}D_{i,g}}.
\end{equation}
To construct an approximate Jacobian $\tilde J(q)$ for a single point we combine the rigid Jacobians with their respective rigidity values
\begin{equation}
    \tilde J(q,i,g) = \begin{bmatrix}w_{trans}(i,g) J_{trans}(q,i,g) & & w_{rot}(i,g) J_{rot}(q,i,g)\end{bmatrix} \enspace ,
\end{equation}
and then combine the results into a single matrix
\begin{equation}
    \tilde J(q) = 
    \begin{bmatrix}
        \tilde J(q,1,1) & \tilde J(q,1,2) & \dots & \tilde J(q, 1, G) \\
        \tilde J(q,2,1) & \ddots \\
        \vdots \\
        \tilde J(q,P,1)
    \end{bmatrix} \enspace .
\end{equation}

\section{Obstacle Proximity Algorithm}
\label{apx:obstacle_proximity}

\begin{algorithm}[H]
    \caption{Proximity$(\gripperindex, \obstacle)$ - reproduced from~\cite{Berenson2013}}
    \label{alg:obstacle_avoidance}
    \begin{algorithmic}[1]
        \State $d_\gripperindex \gets \infty$
        \For{$o \in \{1,2,\dots,|\obstacle\ \}$}
            \State $p^\gripperindex, p^o \gets$ ClosestPoints$(\gripperindex, o)$
            \State $v \gets p^\gripperindex - p^o$
            \If{$\| v \| < d_\gripperindex$}
                \State $d_\gripperindex \gets \| v \|$
                \State $\dot x_{p^\gripperindex} \gets \frac{v}{\| v \|}$
                \State $J_{p^\gripperindex} \gets$ GripperPointJacobian$(\gripperindex, p^\gripperindex)$
            \EndIf
        \EndFor
        \State \Return $\{J_{p^\gripperindex}, x_{p^\gripperindex}, d_\gripperindex \}$
    \end{algorithmic}
\end{algorithm}

\clearpage
\section{Experiment Parameter Values}
\label{apx:param_table}

\begin{table}[ht]
\centering
\caption{Controller parameters}
\label{tab:controller_param_table}
\begin{tabular}{lcccccc}
\hline\noalign{\smallskip}
                                        &                           & \parbox{0.5in}{\centering Synthetic\\Trials} 
                                                                    & \parbox{0.5in}{\centering Rope\\Winding}
                                                                    & \parbox{0.5in}{\centering Table\\Coverage}
                                                                    & \parbox{0.6in}{\centering Two Stage\\Coverage} \\
\noalign{\smallskip}\hline\noalign{\smallskip}
$\tanse{3}$ inner product constant      & $c$                       &   - & 0.0025 & 0.0025 & 0.0025 \\
Servoing max gripper velocity           & $\maxgrippervelservo$     & 0.1 &    0.2 &    0.2 &    0.2 \\
Obstacle avoidance max gripper velocity & $\maxgrippervelobstacle$  &   - &    0.2 &    0.2 &    0.2 \\
Obstacle avoidance scale factor         & $\beta$                   &   - &    200 &   1000 &   1000 \\
Stretching correction scale factor      & $\lambda$                 &   - &  0.005 &   0.03 &   0.03 \\
\hline\noalign{\smallskip}
\end{tabular}
\end{table}

\begin{table}[ht]
\centering
\caption{KF-MANB and KF-MANDB parameters}
\label{tab:param_table}
\begin{tabular}{lcccccc}
\hline\noalign{\smallskip}
                                        &                               & \parbox{0.5in}{\centering Synthetic\\Trials} 
                                                                        & \parbox{0.5in}{\centering Rope\\Winding}
                                                                        & \parbox{0.5in}{\centering Table\\Coverage}
                                                                        & \parbox{0.6in}{\centering Two Stage\\Coverage} \\
\noalign{\smallskip}\hline\noalign{\smallskip}
\parbox{1.5in}{Correlation strength factor\\(KF-MANDB only)}            & $\correlationfactor$          &  0.9 &   0.9 &   0.9 &   0.9 \\
\noalign{\smallskip}
Transition noise factor                                                 & $\transitionnoisefactor^2$    &    1 &   0.1 &   0.1 &   0.1 \\
Observation noise factor                                                & $\observationnoisefactor^2$   &    1 &  0.01 &  0.01 &  0.01 \\
\hline\noalign{\smallskip}
\end{tabular}
\end{table}

\end{document}